\documentclass{article}

    \PassOptionsToPackage{round}{natbib}
    \usepackage[preprint]{neurips_2022}

\usepackage[utf8]{inputenc} 
\usepackage[T1]{fontenc}    
\usepackage{hyperref}       
\usepackage{url}            
\usepackage{booktabs}       
\usepackage{amsfonts}       
\usepackage{nicefrac}       
\usepackage{microtype}      
\usepackage{xcolor}         

\usepackage{enumitem}
\usepackage{graphicx}
\usepackage{changepage}
\usepackage{placeins}
\usepackage{float}
\usepackage{multirow}
\usepackage{caption}
\usepackage{subcaption}
\usepackage{longtable}

\title{Systematic Generalization and Emergent Structures in Transformers Trained on Structured Tasks}

\author{%
  Yuxuan Li \\
  Department of Psychology\\
  Stanford University\\
  Stanford, CA 94305 \\
  \texttt{liyuxuan@stanford.edu} \\
   \And
  James L. McClelland \\
  Department of Psychology\\
  Stanford University\\
  Stanford, CA 94305 \\
  \texttt{jlmcc@stanford.edu} \\
}

\begin{document}

\maketitle

\begin{abstract}
Transformer networks have seen great success in natural language processing and machine vision, where task objectives such as next word prediction and image classification benefit from nuanced context sensitivity across high-dimensional inputs. However, there is an ongoing debate about how and when transformers can acquire highly structured behavior and achieve systematic generalization. Here, we explore how well a causal transformer can perform a set of algorithmic tasks, including copying, sorting, and hierarchical compositions of these operations. We demonstrate strong generalization to sequences longer than those used in training by replacing the standard positional encoding typically used in transformers with labels arbitrarily paired with items in the sequence. We search for the layer and head configuration sufficient to solve these tasks, then probe for signs of systematic processing in latent representations and attention patterns.  We show that two-layer transformers learn reliable solutions to multi-level problems, develop signs of task decomposition, and encode input items in a way that encourages the exploitation of shared computation across related tasks.  These results provide key insights into how attention layers support structured computation both within a task and across multiple tasks.
\end{abstract}

\section{Introduction}

Since their introduction \citep{vaswani2017attention}, transformer-based models have become the new norm of natural language modeling \citep{brown2020language, devlin2018bert} and are being leveraged for machine vision tasks as well as in reinforcement learning contexts \citep{chen2021decision, dosovitskiy2020image, janner2021reinforcement, ramesh2021zero}.  Transformers trained on large amounts of data under simple self-supervised, sequence modeling objectives are capable of subsequent generalization to a wide variety of tasks, making them an appealing option for building multi-modal, multi-task, generalist agents \citep{bommasani2021opportunities, reed2022generalist}.

Central to this success is the ability to represent each part of the input in the context of other parts through the self-attention mechanism.  This may be especially important for task objectives such as next word prediction and image classification at scale with naturalistic data, which benefit from nuanced context sensitivity across high-dimensional inputs.  Interestingly, transformer-based language models seem to also acquire some knowledge of syntactic structures without being explicitly trained to do so and display few-shot learning capabilities \citep{brown2020language, linzen2021syntactic, manning2020emergent}.  These insights have led to ongoing work assessing broader reasoning capabilities in these models \citep{binz2022using, dasgupta2022language}.

Despite success in learning large-scale, naturalistic data and signs of generalizable behavior or sensitivity to structures, how transformer models support systematic generalization remains to be better understood.  Recent work demonstrated that large language models struggle at longer problems and fail to robustly reason beyond the training data \citep{anil2022exploring, razeghi2022impact}.  Different architectural variations have been proposed to improve length generalization in transformers, highlighting the role of variants of position-based encodings \citep{csordas2021devil, csordas2021neural, ontanon2021making, press2021train}.  Indeed, whether neural networks will ever be capable of systematic generalization without building in explicit symbolic components remains an open question \citep{fodor1988connectionism, smolensky2022neurocompositional}.

Here, we approach this question by training a causal transformer to perform a set of algorithmic operations, including copy, reverse, and hierarchical group or sort tasks.  We explicitly sought the minimal transformer that would reliably solve these simple tasks and thoroughly analyze such minimal solution through attention ablation and representation analysis to understand the internal computational dynamics.  Exploring how a transformer with no predefined task-aligned structure could adapt to structures in these algorithmic tasks provides a starting point for understanding how self-attention can tune to structures in more complex problems, e.g., those with the kinds of exceptions and partial regularities of natural datasets, where the exploitation of task structures may occur in a more approximate and graded manner.
Our main contributions are:
\begin{enumerate}
    \item We highlight a simple label-based order encoding method in place of the positional encoding methods typically used in transformers, and show that it helps our models achieve strong length generalization performance across the set of algorithmic tasks we examine.
    \item We thoroughly analyze simple, two-layer causal transformers that learn these algorithmic tasks, and show that the attention layers develop signs of systematic decomposition within tasks and exploitation of shared structures across tasks.
\end{enumerate}

\section{Method}

\begin{figure}[h]
    \centering
    \includegraphics[width=14cm]{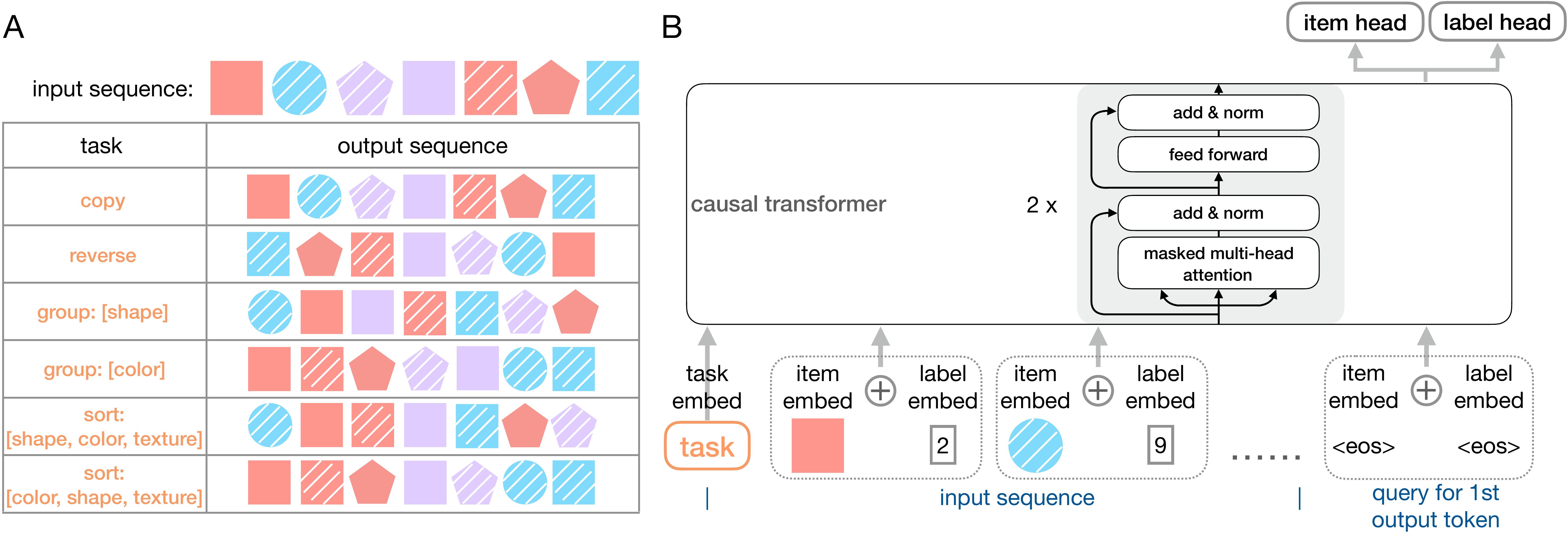}
    \caption{Task and model design.}
    \label{fig:modeltask}
\end{figure}

\textbf{Dataset}.  
We created an item pool covering all combinations of 5 shapes, 5 colors, and 5 textures, and generated a sequence dataset by sampling 100k sequences of 5--50 items randomly selected from the item pool.  The tasks we used to train the models are shown in Fig~\ref{fig:modeltask}A.  Each task corresponds to one of the following rules, which relies on item feature and/or item order information to rearrange an input sequence (grouping or sorting items by a particular feature is with respect to a pre-defined feature sort order, e.g., circles $<$ squares $<$ pentagons, or red $<$ purple $<$ blue):
\begin{description}[noitemsep]
    \item \textsc{copy (C)}: copy the input sequence.
    \item \textsc{reverse (R)}: reverse the input sequence.
    \item \textsc{group[shape] (G[s])}: group the items by shape, preserve the input order within each shape group.
    \item \textsc{group[color] (G[c])}: group the items by color, preserve the input order within each color group.
    \item \textsc{sort[shape,color,texture] (S[s])}: sort the items first by shape, then by color, then by texture.
    \item \textsc{sort[color,shape,texture] (S[c])}: sort the items first by color, then by shape, then by texture.
\end{description}

We instantiated the token vocabularies as onehot or multihot vectors.  The task tokens were onehot vectors with the corresponding task category set to one, with one additional task dimension corresponding to the end-of-sequence (\textsc{eos}) token.  The item tokens were multihot vectors whose units indicated its value in each feature dimension (equivalent to concatenated onehot feature vectors).  As such, the model receives disentangled feature information in the input, though in principle it can learn to disentangle feature information given onehot encodings for each unique item.

\textbf{Label-based order encoding}.
Using position-based order encodings, models trained with sequences up to length $L$ encounter an out-of-distribution problem when tested on longer sequences, as position encodings beyond $L$ are unfamiliar to the model.  We introduce label-based encoding, which instead pairs items in each sequence with ascending random integer labels to communicate order information  (Fig~\ref{fig:modeltask}B).  This allows models to encode longer sequences of tokens with familiar labels seen during training.  In our model, these labels were embedded with learnable weights, and we contrast the random label encoding method with sinusoidal and learnable encodings based on item positions.  A concurrent work also explored the random position method and tested with other types of encodings \citep{acl2022randomized}.  In all reported results, we pre-generated item labels sampled from a range up to the maximum generalization length (50) for all sequences in the dataset, and these labels were shared across training steps and model seeds.  In practice, the labels for each sequence can be sampled online and from a larger range to accommodate generalization to even longer sequences.

\textbf{Model}.
The main model architecture is shown in Fig~\ref{fig:modeltask}B.  Each input sequence consisted of a task token and the paired item and label tokens, with the \textsc{eos} token serving as the first query for tokens in the output sequence.  The input tokens were first embedded to the model's latent representational space through a set of embedding layers depending on the token type (task, item, or label).  The item and label embeddings were then added to form a composite item embedding.  These embedded tokens were fed into a causal transformer, which contained one or two layers of alternating future-masked attention sublayers and MLP sublayers.  Residual connections and layer normalization were applied after each sublayer as in \citet{vaswani2017attention}.   We tested architectural variations in the number of attention heads in different layers of the model while controlling for the total number of learnable parameters (see detailed hyperparameters in Appendix~\ref{sec:hyperparam}).  The state of the query token at the output of the causal transformer was passed through two linear heads to predict the next output token (the task token, or an item and its associated label).

\textbf{Training and evaluation}.  
The models were trained using full teacher forcing (where we always feed the model the correct tokens) on all sequences of lengths 5 to 25 in the dataset ($\sim$46k) and evaluated for length generalization on sequences of lengths 26 to 50 ($\sim$54k).  We trained models in both single-task and multi-task settings.  In both cases, the output sequence consisted of the correctly ordered items and their labels given the task being trained, followed by an \textsc{eos} token.  In single-task learning, we did not include the task token in training or evaluation.  In multi-task learning, the task token was used and the models were trained to first output the task token before predicting the output sequence.  The training sequences used in multi-task learning remained the same ones between lengths 5--25, but each sequence corresponded to a different output sequence under different tasks.  

The models were trained using softmax cross-entropy loss on the prediction of feature classes, labels, and task/\textsc{eos} categories for tokens in the output sequence.  Item predictions were treated as average feature prediction accuracy, i.e., if the model predicted 2/3 features correct, its token-level item accuracy is 2/3.  Training stopped at 32k gradient updates for single-task models and 38k gradient updates for multi-task models.  Below, we report both token-level and sequence-level accuracy, under both teacher forcing and top1 rollout (i.e., greedy decoding). Results were aggregated over four random seeds for each task type $\times$ architecture pair.  Unless otherwise specified, results were taken from the checkpoint with the highest generalization accuracy within each seed.  Error shades and error bars indicate standard error of the mean across models.

\section{Results}
\subsection{Single-task learning}

\begin{figure}[h]
    \centering
    \includegraphics[width=14cm]{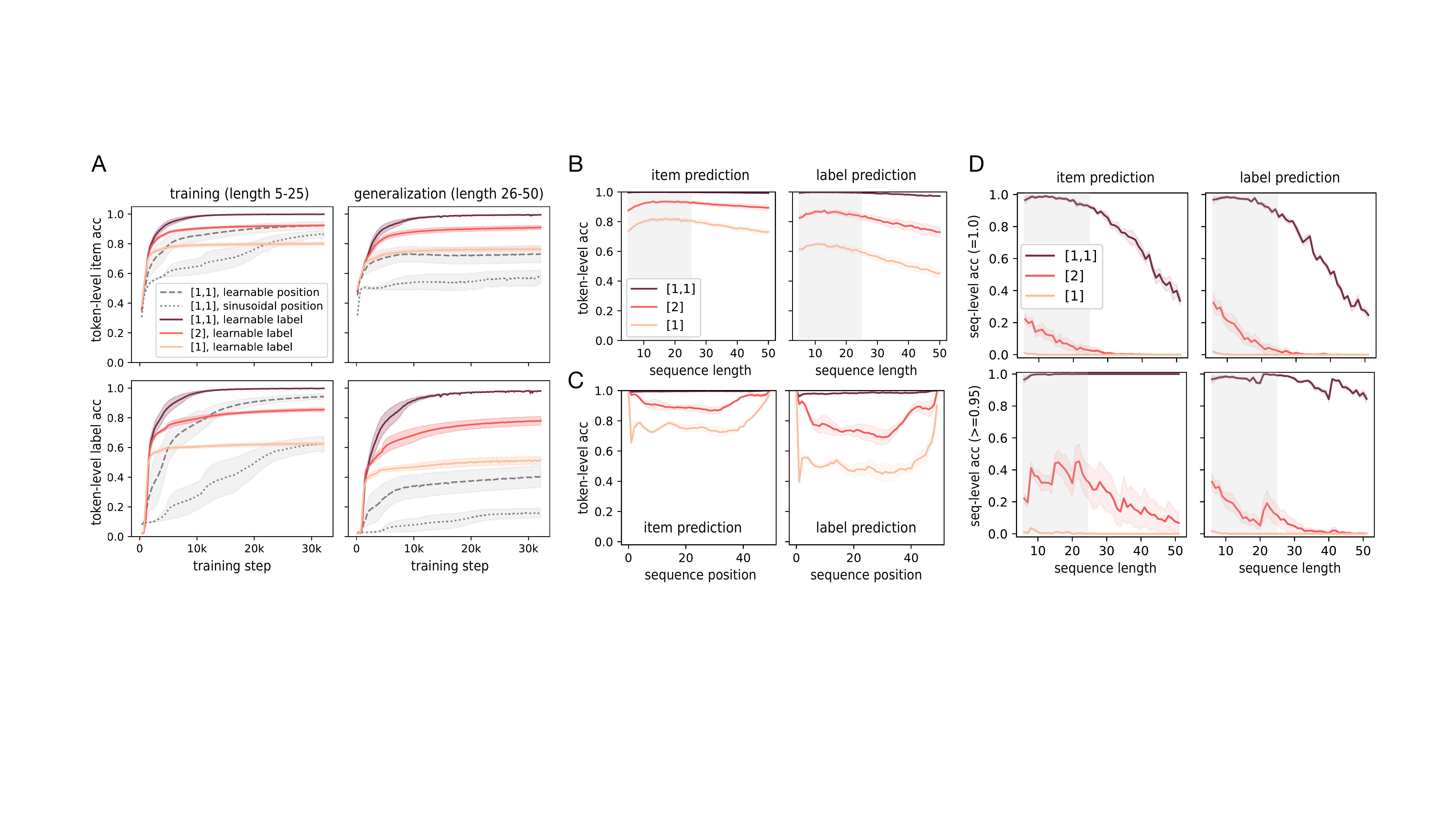}
    \caption{
    Token- and sequence-level accuracy for single-task models.
    \textbf{A}. Token-level accuracy on training and generalization sequences over learning.
    \textbf{B}. Token-level accuracy over sequence length.
    \textbf{C}. Token-level accuracy over sequence positions.
    \textbf{D}. Proportion of sequences that the model predicted 100\% tokens correct (upper) or predicted greater than 95\% tokens correct (lower).  In B, C, and D, results were taken from 5k novel sequences in the training length range (in B and D) and 5k generalization sequences (B, C, and D).  Legends indicate the number of attention heads in each layer and the order encoding used (in A).  Gray shades indicate the range of lengths used in training.
    }
    \label{fig:single-task-performance}
\end{figure}

\textbf{Two-layer models with label encoding learn the \textsc{sort} task and generalize to longer sequences}. We first trained the model with the \textsc{sort[shape,color,texture]} task.  Using our label encoding method, models with two single-headed layers (indicated as [1,1]) were able to achieve near-ceiling accuracy on training sequences and generalize to longer sequences (Fig~\ref{fig:single-task-performance}; also see quantitative results in Appendix~\ref{sec:quant-result}).  The predictions of the \textsc{eos} token were also highly accurate in these models (see Fig~\ref{fig:st-append-tf}A in Appendix~\ref{sec:st-append}).  Item prediction was more accurate than label prediction in this task, reflecting that the models represented item feature information more accurately in order to sort the input tokens.  The two-layer models showed some degradation in sequence-level accuracy as a function of sequence length, but the failures on longer sequences were not catastrophic, as these models scored very well on longer sequences when up to 5\% prediction errors were allowed (Fig~\ref{fig:single-task-performance}D; also see Fig~\ref{fig:st-append-tf}B, and Fig~\ref{fig:st-append-ro} for accuracy under rollout in Appendix~\ref{sec:st-append}).  In contrast, two-layer models trained with sinusoidal or learnable position encodings performed worse across both training and generalization sequences (Fig~\ref{fig:single-task-performance}A).

The two-layer models were also much better than single-layer models with either one or two attention heads.  While these single-layer models were able to exploit some correlations between items and output positions (e.g., item [0,0,0] always came first, and item [4,4,4] always came last), they failed to sort items in the middle positions (Fig~\ref{fig:single-task-performance}C).  In contrast, a single-layer, single-headed model was sufficient to learn the \textsc{copy} or the \textsc{reverse} task (see Fig~\ref{fig:st-append-other}A in Appendix~\ref{sec:st-append}), suggesting that multiple layers strongly benefit successful learning of multi-level problems.

\begin{figure}[h]
    \centering
    \includegraphics[width=14cm]{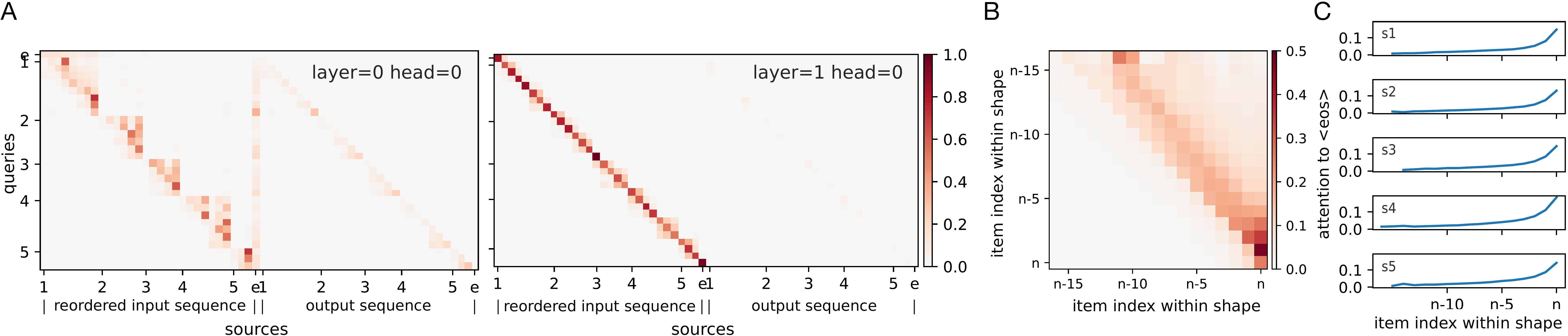}
    \caption{Attention patterns in two-layer models.
    \textbf{A}. Attention maps for an example generalization sequence.  Items in the input sequence were reordered to match their output order for visualization purposes.  Numbers 1-5 mark the beginning of each shape group.  Label e indicates the \textsc{eos} token.
    \textbf{B}. First-layer attention from query items to source items within shape groups.  
    \textbf{C}. Attention to \textsc{eos} as a function of item index within each shape group (indicated by labels s1-s5).
    Results in B and C were aggregated across 1k generalization sequences and across seeds.
    }
    \label{fig:single-task-attn}
\end{figure}

\textbf{Distinct two-stage processing across attention layers}.  The attention weights in the two-layer models revealed signs of task decomposition (Fig~\ref{fig:single-task-attn}A).  The attention head in the first layer tended to distribute attention to the unsorted items that share the same shape as the current query item.  The attention head in the second layer then almost exclusively attended to the next output token in preparation for feature and label readout.  This pattern appeared robustly across sequences and across different seeds (Fig~\ref{fig:single-task-attn}B).  Interestingly, there was an increase in the attention weights to the \textsc{eos} token as the model received query items towards the end of each shape group. This attention to \textsc{eos} increased to similar degrees in early or late shape groups (Fig~\ref{fig:single-task-attn}C), again suggesting that the model learned to systematically process items within each shape group, even though generating the \textsc{eos} token was only relevant after sorting all items.  We also found similar attention patterns in two-layer, single-headed models learning the \textsc{group[shape]} task (see Fig~\ref{fig:st-append-other}B in Appendix~\ref{sec:st-append}).

The single-layer models displayed some attention to subsequent items in the output sequence but lacked consistent structures across different shape groups (see Fig~\ref{fig:st-append-attn} in Appendix~\ref{sec:st-append}).  This could be due to the burden for the attention head(s) within a single layer to implement a mixture of item contextualization and readout of the correct item or label, and reflects an advantage of the two-layer models in injecting an inductive bias for a multi-stage solution.

\begin{figure}[h]
    \centering
    \includegraphics[width=9.5cm]{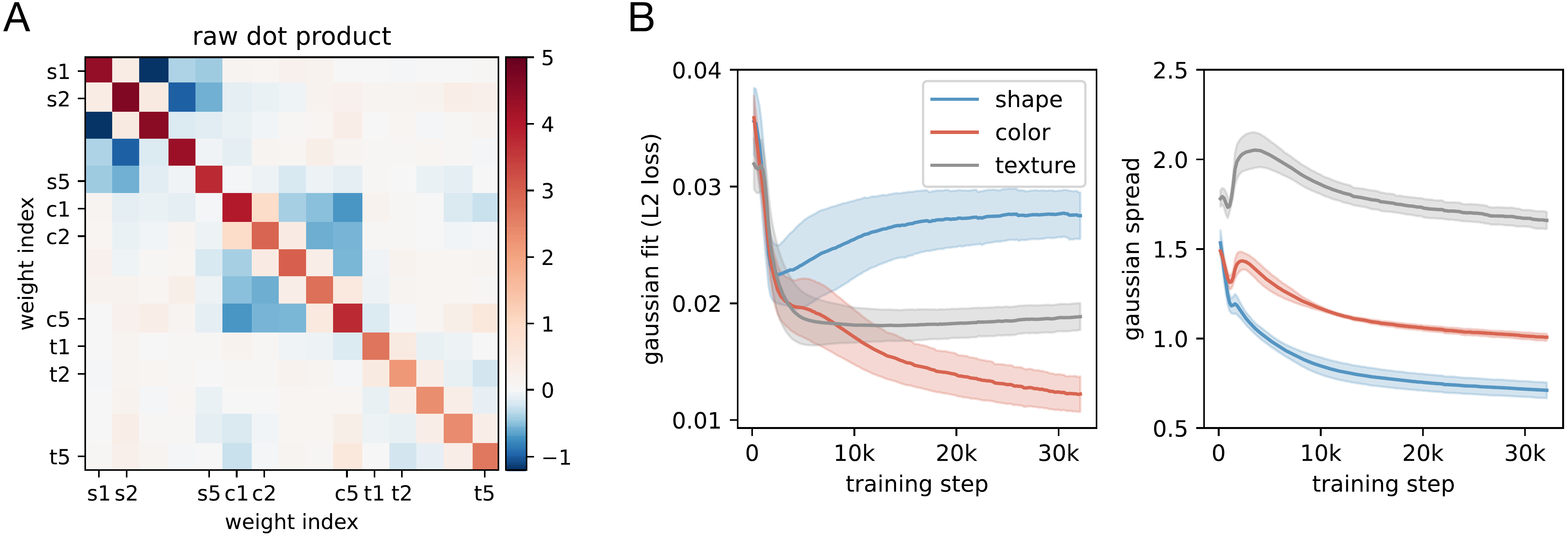}
    \caption{Input item embeddings capture information about feature sort order.
    \textbf{A}. Dot product similarity measure of item embedding weights. s1-s5/c1-c5/t1-t5: shape/color/texture embedding weights.
    \textbf{B}. Fitting Gaussian representations to track learned feature sort order in the input embeddings. Left, fit objective (mean L2 loss) over pairwise similarities of feature embeddings.  Right, best-fitting Gaussian spread parameter for each feature.
    }
    \label{fig:single-task-order}
\end{figure}

\textbf{Acquisition of within-feature order information}.  To solve the sort task accurately, the models needed to learn the invariant sort order within each feature.  We tested if the learned order information can be parsed out from the input embeddings.  We found that the embedding weights associated with shape and color feature units reflected within-feature order similarities, with weights associated with different features appearing mostly orthogonal (Fig~\ref{fig:single-task-order}A).  The weaker structure and smaller magnitude associated with texture weights may be due to lower demand in sorting multiple texture values, as texture was the third-level sort feature and thus had fewer consecutive values in a single output sequence under limited sequence length.

We quantitatively tracked how information about feature sort order was acquired in the input embeddings over learning, using an L2 loss over the difference in the pairwise similarities from synthetic Gaussian feature representations and that from the models' feature embeddings.  This analysis suggested that the two-layer models initially began to acquire feature sort order information concurrently in all three features (Fig~\ref{fig:single-task-order}B, left panel).  In later stages during learning, the sort order information in color embeddings more quickly and better corresponded to similarities between Gaussian representations, with shape embeddings deviating from Gaussian-like, monotonic order similarities.

\subsection{Multi-task learning}

\textbf{Multi-task learning and length generalization in two-layer, multi-headed models}. To explore the ability for the causal transformer to simultaneously learn multiple algorithmic tasks, we trained models to predict different output sequences on the same input sequence conditioned on a task token.  The two-layer, single-headed model used in single-task learning was unable to learn all six tasks, while two-layer, multi-headed models achieved good training and generalization performance (Fig~\ref{fig:multi-task-acc}A, also see quantitative performance in Appendix~\ref{sec:quant-result}).  Increasing the number of attention heads in the model did not lead to drastically different learning curves under teacher forcing, but more attention heads supported much better performance under top1 rollout (Fig~\ref{fig:multi-task-acc}B).

We tested whether multi-headed attention served for better learning when it occurred in the first layer (attention-frontload) or the second layer (attention-backload). Attention-backload models generally achieved higher accuracy across all tasks compared to their attention-frontload counterparts (Fig~\ref{fig:multi-task-acc}C and \ref{fig:multi-task-acc}D).  Performance of the attention-backload models was even comparable with their attention-balanced counterparts, which signals that a bottlenecked architecture may be particularly suited for multi-task learning in our setting, considering that some tasks in the task suite share the first-level grouping feature.

\begin{figure}[h]
    \centering
    \includegraphics[width=14cm]{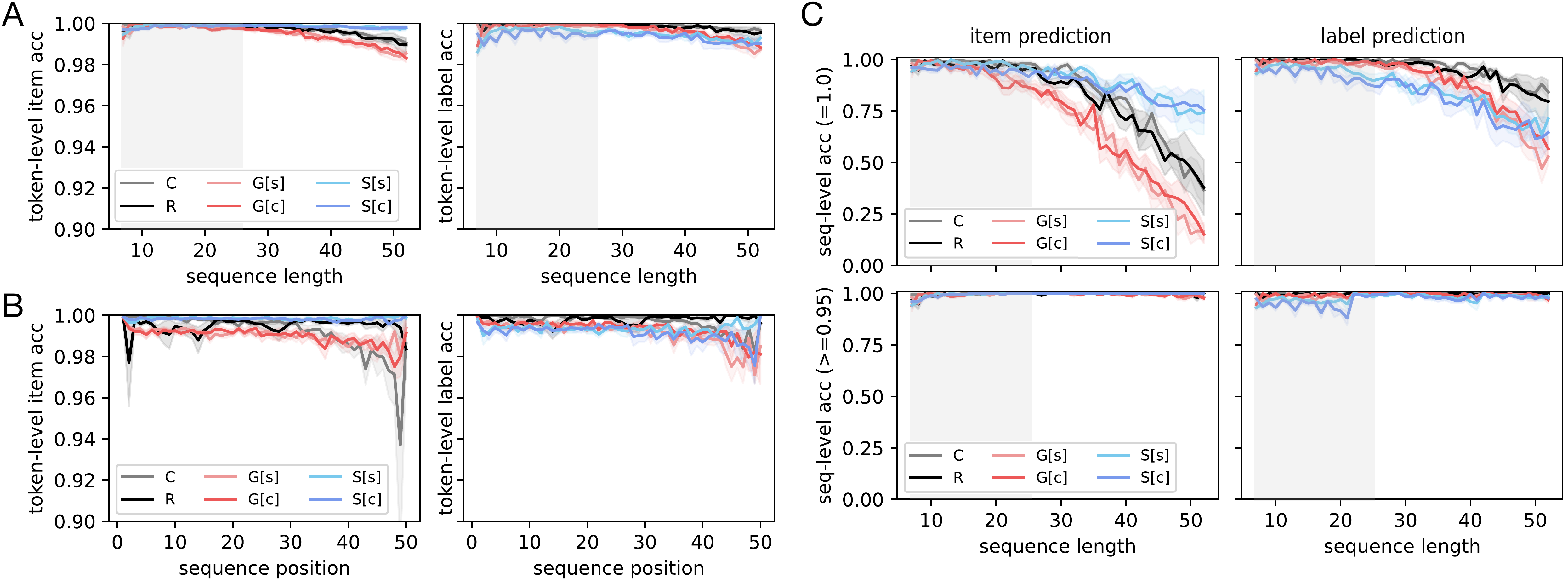}
    \caption{
    Length generalization across tasks.
    \textbf{A} and \textbf{B}. Token-level item and label accuracy over sequence length and sequence position.
    \textbf{C}. Proportion of sequences that the model predicted 100\% tokens correct (upper) or predicted greater than 95\% tokens correct (lower).
    Results were taken from 1k novel sequences in the training length range (in A and C) and 1k generalization sequences (in A, B, and C) for each task.  Performances were aggregated across the five top-performing runs.  Gray shades indicate the range of lengths used in training.
    }
    \label{fig:multi-task-len}
\end{figure}

Consistent with the single-task model, six-task models demonstrated strong generalization to longer sequences (Fig~\ref{fig:multi-task-len}A).  Even though sequence-level accuracy degraded as sequence length increased, the models only made less than 5\% prediction errors for long extrapolation sequences (Fig~\ref{fig:multi-task-len}C).  Token-level accuracy among generalization sequences was also relatively stable until the last few output positions (Fig~\ref{fig:multi-task-len}B).  We also observed that accurately predicting item features in long sequences was easier in the \textsc{sort} tasks but harder in the \textsc{copy}, \textsc{reverse}, and \textsc{group} tasks (Fig~\ref{fig:multi-task-len}A and \ref{fig:multi-task-len}C).  This echos results from the single-task models and again suggests that the models more accurately represented information directly used for sorting the items in each task.  See additional results on \textsc{eos} prediction and accuracy under rollout in Appendix~\ref{sec:mt-append}.

\begin{figure}[h]
    \centering
    \includegraphics[width=14cm]{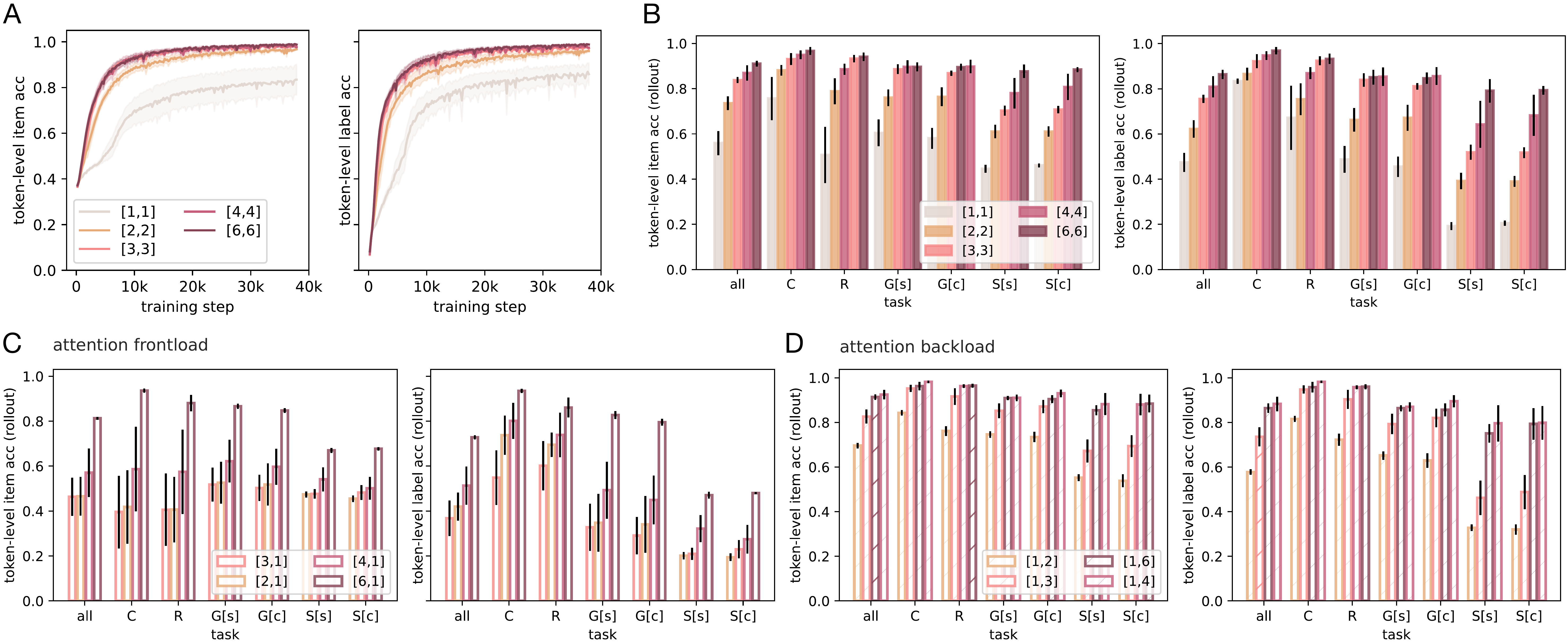}
    \caption{
    Token-level item and label accuracy in multi-task models.  
    Legends indicate the number of attention heads in the two layers.
    \textbf{A}.  Accuracy across all tasks over learning.
    \textbf{B}, \textbf{C}, and \textbf{D}. Top1 rollout accuracy for models with attention-balanced, attention-frontload, and attention-backload architectures.
    In A, results were taken from 75k generalization sequences (12.5k per task).  In B, C, and D, results were taken from 1.2k generalization sequences (200 per task).
    }
    \label{fig:multi-task-acc}
\end{figure}

\begin{figure}[h]
    \centering
    \includegraphics[width=14cm]{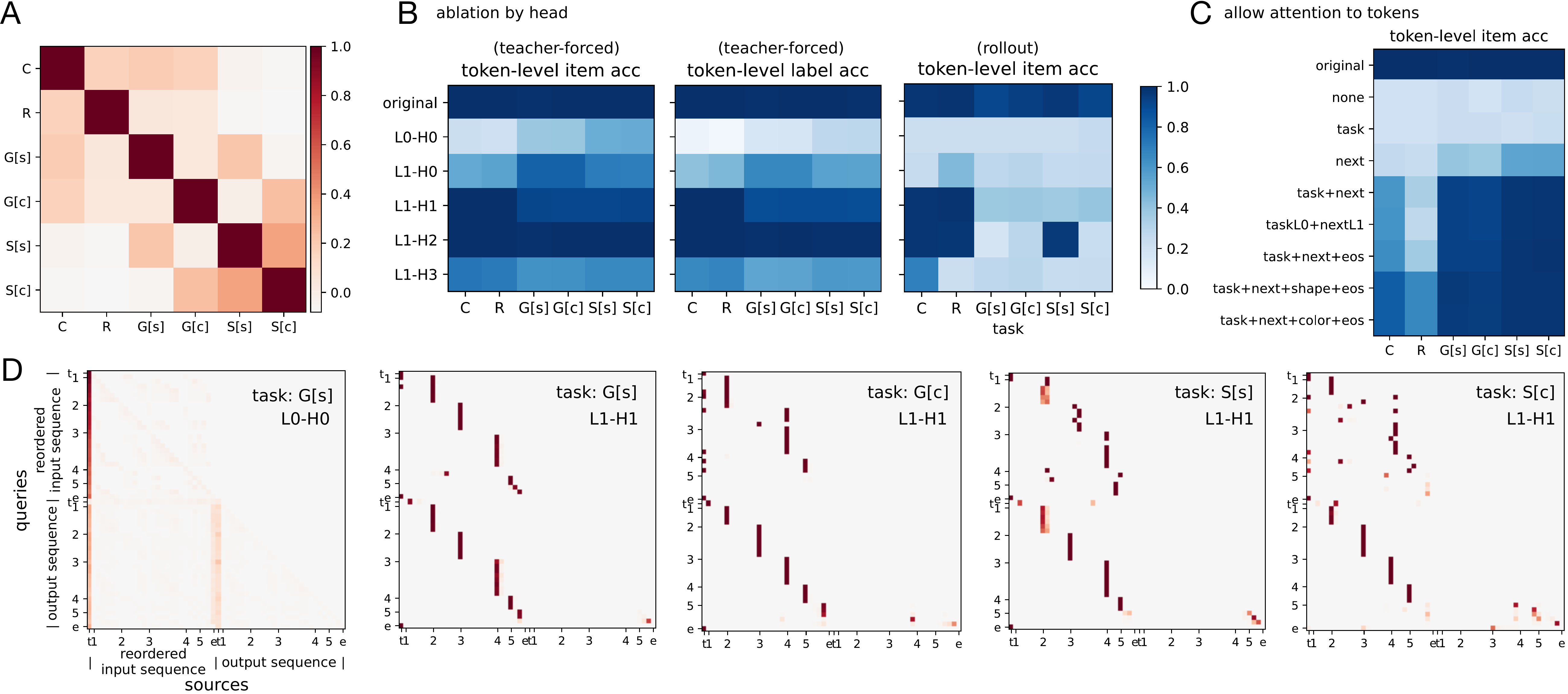}
    \caption{
    \textbf{A}. Pairwise similarities of task embeddings.
    \textbf{B}. Token-level accuracy from ablating single attention heads.
    \textbf{C}. Token-level accuracy from preserving attention to certain tokens across attention heads.
    \textbf{D}.  Attention maps from selected attention heads in an example generalization sequence (see Fig~\ref{fig:mt-append-attn-maps} for all attention maps).   Items corresponding to the input sequence were reordered to match their output order for visualization purposes. Numbers 1-5 mark the beginning of each first-level feature group (shape or color). Label t indicates the task token.  Label e indicates the \textsc{eos} token.
    In B, C, and D, results were taken from the top-performing model (architecture: [1,4]).
    }
    \label{fig:multi-task-embed-attn}
\end{figure}

\textbf{Learned task embeddings recover task similarity.}  We examined the input task embeddings to understand the basis of task-shared and task-specific computation in the model.  Fig~\ref{fig:multi-task-embed-attn}A shows that similarities among the learned task embeddings reflected similarities across tasks.  For example, the \textsc{copy} task was recognized to be similar to the \textsc{reverse} task and the two \textsc{group} tasks, potentially reflecting the shared need for stronger representation of label information in order to accurately sort the input items.  Representations for the two \textsc{sort} tasks were also highly similar, as they both rely on item features to sort the input items.  The models also recovered similarities between the pairs of \textsc{group} and \textsc{sort} tasks that share the same first-level grouping feature.

\textbf{Task-shared and task-dependent computation across attention layers.}  Structures in the task embedding similarities and strong performance from models with fewer attention heads than there are tasks already hinted that the model could be exploiting shared processing across tasks.  Because task-conditioned computation can only occur across the attention layers, we next sought to understand the role of multi-headed attention in implementing such shared computation.  We performed two ablation experiments: ablating a single attention head entirely and preserving attention in all other heads, or preserving attention to certain tokens across all attention heads.  Attention weights were ablated by masking the attention weights as zero after softmax.

Fig~\ref{fig:multi-task-embed-attn}B and C show the ablation results for the top-performing model with one attention head in the first layer and four attention heads in the second layer.  The attention heads did not exhibit strong selectivity for single tasks as they usually contributed to multiple tasks, and they also showed equal contribution to item and label prediction (Fig~\ref{fig:multi-task-embed-attn}B, also see results from other models in Fig~\ref{fig:mt-append-attn-pca} in Appendix~\ref{sec:mt-append}).  One attention head appeared redundant as ablating it resulted in little impact on the performance of any task under teacher forcing, but it significantly improved accuracy under rollout.  

Ablating attention to certain items in the sequence further indicated that the models were heavily relying on the learnable task embedding to contextualize items under different tasks (Fig~\ref{fig:multi-task-embed-attn}C, also see Fig~\ref{fig:mt-append-attn-pca}).  When the models were only allowed to attend to the task token in the first layer and the next output token in the second layer, performance was largely preserved in the \textsc{group} tasks and the \textsc{sort} tasks (ablation type ``taskL0+nextL1'').  This is different from the two-layer single-task models, whose first layer relied on attention to other items in the first-level feature group to cross-contextualize items.  However, there were still some signs of task decomposition in the multi-task model.  For example, one attention head in the second layer consistently attended to the first item in the next feature group across all multi-level tasks (Fig~\ref{fig:multi-task-embed-attn}D; see full attention maps in Fig~\ref{fig:mt-append-attn-maps} in Appendix~\ref{sec:mt-append}), which proved useful for accurate predictions across multi-level tasks under both teacher forcing and rollout (Fig~\ref{fig:multi-task-embed-attn}B).

\begin{figure}[h]
    \centering
    \includegraphics[width=14cm]{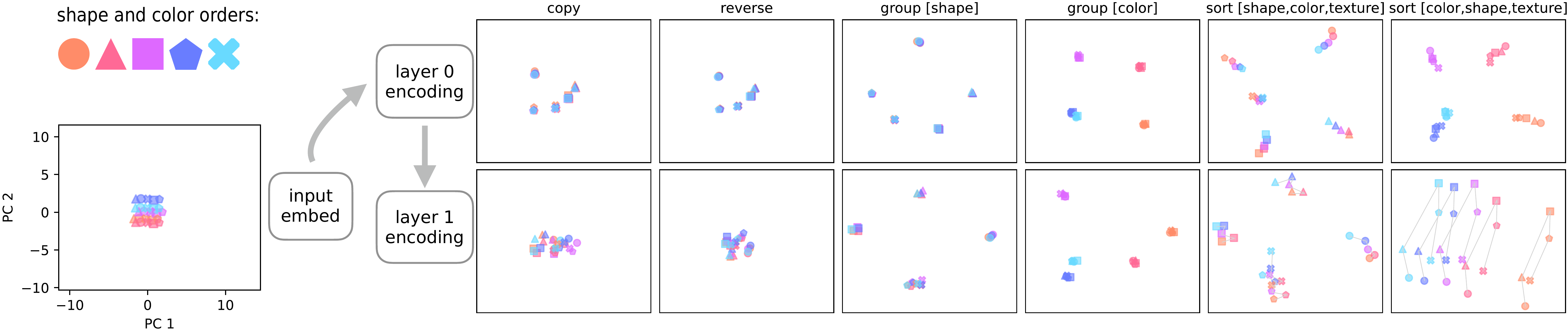}
    \caption{
    Task-conditioned encoding of the same sequence of items from the top-performing model.  PCA was performed for average item representations across 200 generalization sequences within each layer $\times$ task pair.  The legend indicates the predefined sort order of the shape and color features.
    }
    \label{fig:pca}
\end{figure}

\textbf{Task-conditioned item encoding.}  To better understand task-conditioned item representations, we analyzed how the items were encoded under different tasks after each attention layer.  We computed average item representations of all 25 shape $\times$ color feature pairs in each task, aggregating across texture and label values in 200 generalization sequences.  Projecting these item representations to lower dimensions revealed that the model reorganizes representations of item features depending on the task structure (Fig~\ref{fig:pca}).  For the two \textsc{group} tasks, representations in the first layer encoded all items by their relevant first-level grouping feature (shape or color). The item representations did not consistently encode other feature information across both layers, presumably because label information became more relevant for sorting items within each feature group.

Interestingly, for the two \textsc{sort} tasks, the encoded representations after the first layer also showed a high degree of clustering in the corresponding first-level grouping feature, with the secondary sort feature represented in a ring structure.  The subsequent layer then expanded the item representations along the secondary feature in a way that was consistent across all first-level feature groups.  Notably, the clustering effect and the ring structure in the first-layer representations in the \textsc{sort} tasks were more strongly observed in the attention-backload models (see Fig~\ref{fig:mt-append-attn-pca} in Appendix~\ref{sec:mt-append}).  This could suggest that room for flexible multi-headed transformation downstream can relax first-layer encoding, enabling a more consistent solution to emerge across the \textsc{group} and \textsc{sort} tasks.

\section{Related Work}

There is a growing interest in analyzing small models  in more controlled task settings to better understand the capabilities and detailed computations in transformers.  For example, \cite{power2022grokking} explored learning and generalization dynamics in two-layer causal transformers learning binary operations, and \cite{elhage2021mathematical} explored mechanistic interpretability in one- and two-layer transformers without MLP sublayers.  Our work contributes to these efforts in beginning to shape some understanding of the computation and representation dynamics in transformers using detailed analyses on small-scale models. 

Recent work examining systematic generalization in transformers or pre-trained language models highlighted that length generalization remains a challenge and observed that positional encoding can have a significant impact on the extent to which models can systematically generalize \citep{anil2022exploring, csordas2021devil, deletang2022neural, ontanon2021making}.  Many types of architectural modifications have also been proposed to help transformers achieve better length generalization, including different ways to represent positional information \citep[e.g.][]{csordas2021neural, dehghani2018universal, press2021train, su2021roformer}.  Our label-based encoding method adds to this effort by demonstrating the potential in formulating sequence modeling tasks with a more general item-label binding approach rather than item-position binding.  Concurrent to our work, \citet{acl2022randomized} developed randomized position encoding (equivalent to our label-based encoding method) and showed its advantage over a variety of position-based encodings.

Outside of the context of transformers and language models, the use of synthetic, algorithmic tasks has enabled much understanding of the core capabilities of many models \cite[e.g.][]{graves2014neural}. There has also been some interest in performing algorithmic reasoning with neural networks for its own sake \citep{velivckovic2021neural}.  Correspondingly, \citet{velivckovic2022clrs} recently proposed a benchmark for algorithmic reasoning, and evaluated a variety of graph neural network architectures, all of which struggled to extrapolate algorithms to longer or larger inputs.  Our work shows that self-attention has the potential to adapt to structures in the sequence and find reliable solutions to algorithmic tasks.

Ablation experiments and a variety of representation analyses have been applied to understand the role of the attention mechanism in NLP tasks as well as in transformer-based vision models \citep{chefer2021transformer, manning2020emergent, michel2019sixteen, voita2019analyzing}.  However, it is often debated to what extent attention weights afford model interpretability in these settings \citep{jain2019attention, wiegreffe2019attention, vashishth2019attention}, especially considering head redundancy and the difficulty in correctly attributing relevance over high-dimensional inputs.  We show that at least in simple settings, the attention heads can exhibit some level of interpretability consistent with known task structures.  Similar methods have also been applied to understand unit-level and layer-level dynamics that support multi-task computation in small-scale RNNs \citep{driscoll2022flexible, yang2019task}, which revealed some patterns that are consistent with our findings here, as we discuss below.

\section{Discussion}

We sought to understand how transformers can solve a set of highly-structured algorithmic tasks and systematically generalize.  We presented two-layer causal transformers that can learn copying, reversing, and hierarchical sorting operations that generalize to sequences longer than seen during training.  We found that these models learned to exploit structures within tasks or shared across related tasks and exhibited interesting signatures of task decomposition.  Specifically, the attention layers learned to represent item features in a way that helps subsequent individual attention heads multitask or implement similar computations across item groups.

We highlight that the label-based order encoding method was key to enabling our models to generalize the learned tasks to longer sequences.  The key insight is to sample random labels to communicate sequence order information instead of relying on sequence positions.  This simple extension exposes models to a large range of possible labels evenly during training so that longer sequences can be encoded with familiar labels, and is shown to be effective both in our tasks and in a range of different algorithmic tasks \citep{acl2022randomized}.  Compared to this approach, learnable position encoding not only resulted in poor length generalization, but also slower learning on the training sequences in our tasks -- potentially because the models had fewer sequences to learn to encode later positions compared to early ones.  Sinusoidal position encoding has been noted to have limited length generalization capabilities \citep{ontanon2021making, csordas2021devil}.  In our setting, it may have additionally suffered from the demand to read out the corresponding item positions, which is not common across NLP tasks.  It is worth noting that label-based encoding alone would not represent true item distance information. However, in light of recent work showing that transformers can learn positional information without explicit positional encodings \citep{haviv2022transformer}, it is possible that the benefits of label-based order encoding may transfer to natural language inputs.

Our findings on the contributions of individual attention heads in a multi-headed layer in solving these algorithmic tasks echo results from analyses of language models.  For example, single attention heads are rarely responsible for a particular task or syntactic relationship and can often appear redundant \citep{manning2020emergent, michel2019sixteen, voita2019analyzing}.  We do see some degree of selectivity, with conceptually distinct task components shared across a subset of attention heads.  Recent work studying multi-task computation in RNNs has similarly found that multi-task learning led to the exploitation of reusable computation across related tasks \citep{driscoll2022flexible, yang2019task}.  Interestingly, these recurrent models also exhibited high degrees of task-selectivity at the level of individual units in the hidden layer.  The degree of task-selectivity in different architectural components may vary depending on the inductive biases of different model families and the tasks being learned.  Future work is needed to fully characterize the degree of possible computational modularity in multi-headed attention in relation to task structures.

The flexibility to learn and perform multiple tasks is a key desired capability for machine learning.  Our work here provides insights into the dynamics of within-task and cross-task computations that stacks of attention layers develop when learning highly-structured sequences.  Recent work has explored multi-task learning in transformers at scale and achieved impressive results \citep{lee2022multi, reed2022generalist}.  As transformers are increasingly being leveraged for multi-task and multi-modal learning in domains with richer task structures, it is possible that these models may implicitly learn to decompose complex decisions into reusable, multi-level policies.  In future work, we hope to explore these learning and generalization dynamics in transformer-based agents to understand the acquisition of task-conditioned, multi-level behavioral policies in structured environments.

\subsubsection*{Acknowledgments}
We would like to thank Andrew Nam, Mengye Ren, and members of the Stanford PDP lab for useful discussions, and Andrew Lampinen for comments on the manuscript draft.


\newpage

\appendix
\renewcommand{\thefigure}{S\arabic{figure}}
\setcounter{figure}{0}
\renewcommand{\thetable}{S\arabic{table}}
\setcounter{table}{0}

\section{Additional Results}

\subsection{Single-task learning}
\label{sec:st-append}

\begin{figure}[h]
    \centering
    \includegraphics[width=10cm]{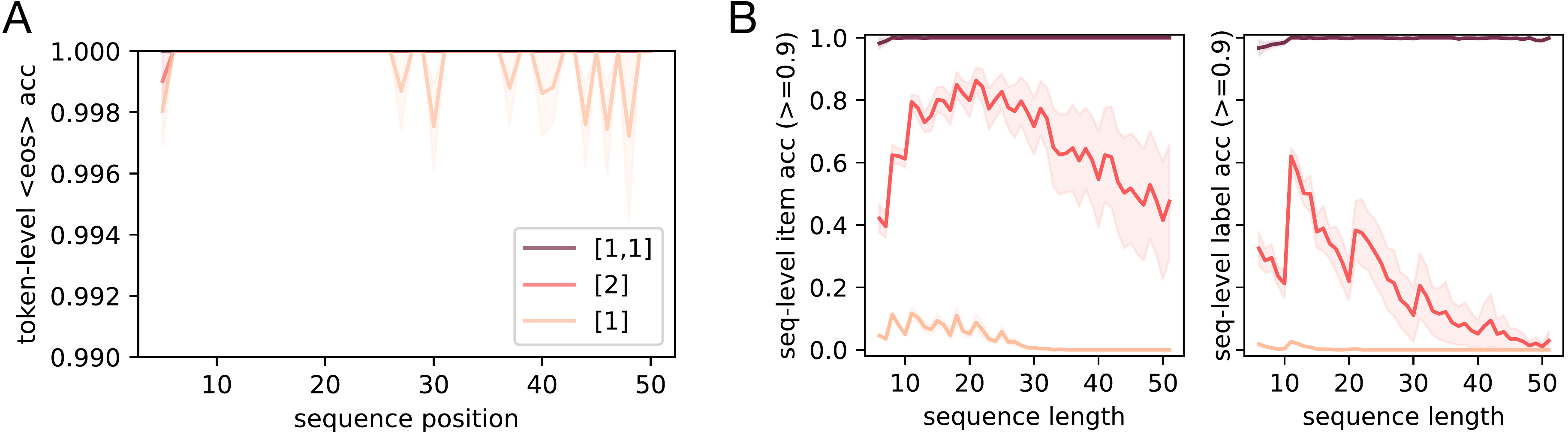}
    \caption{ 
    Additional teacher forcing results.
    \textbf{A}. Single-task models predict \textsc{eos} token near perfectly.
    \textbf{B}. Sequence-level accuracy when up to 10\% errors were allowed.
    Results were taken from 5k novel sequences in the training length range (in B) and 5k generalization sequences (in A and B).
    }
    \label{fig:st-append-tf}
\end{figure}

\begin{figure}[h]
    \centering
    \includegraphics[width=10cm]{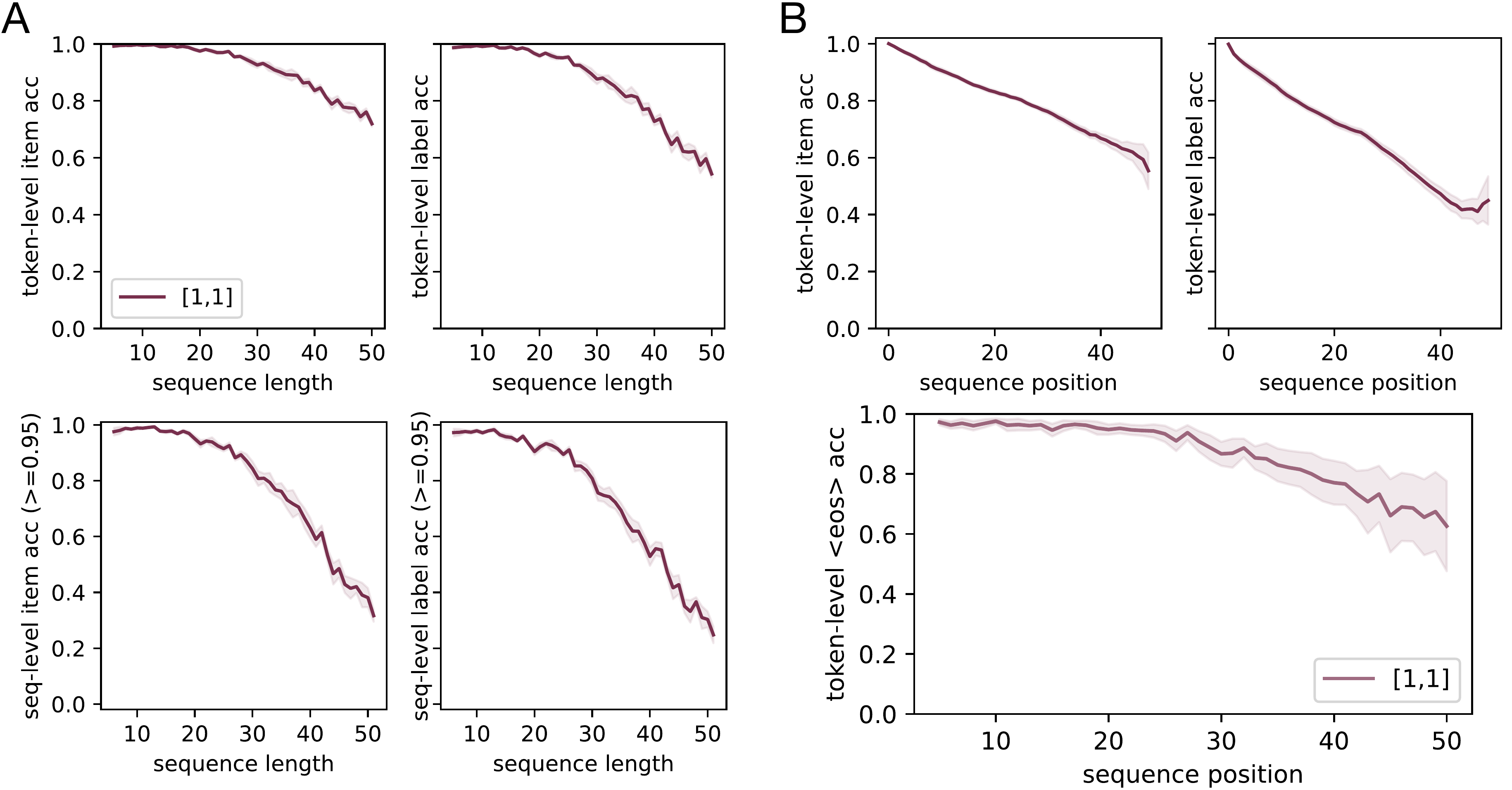}
    \caption{ 
    Top1 rollout accuracy for two-layer models.
    \textbf{A}. Token-level (upper) and sequence-level (lower) accuracy across sequence length.
    \textbf{B}. Token-level accuracy across sequence positions.
    Results were taken from 5k novel sequences in the training length range (in A) and 5k generalization sequences (in A and B).
    }
    \label{fig:st-append-ro}
\end{figure}

\begin{figure}[h]
    \centering
    \includegraphics[width=12cm]{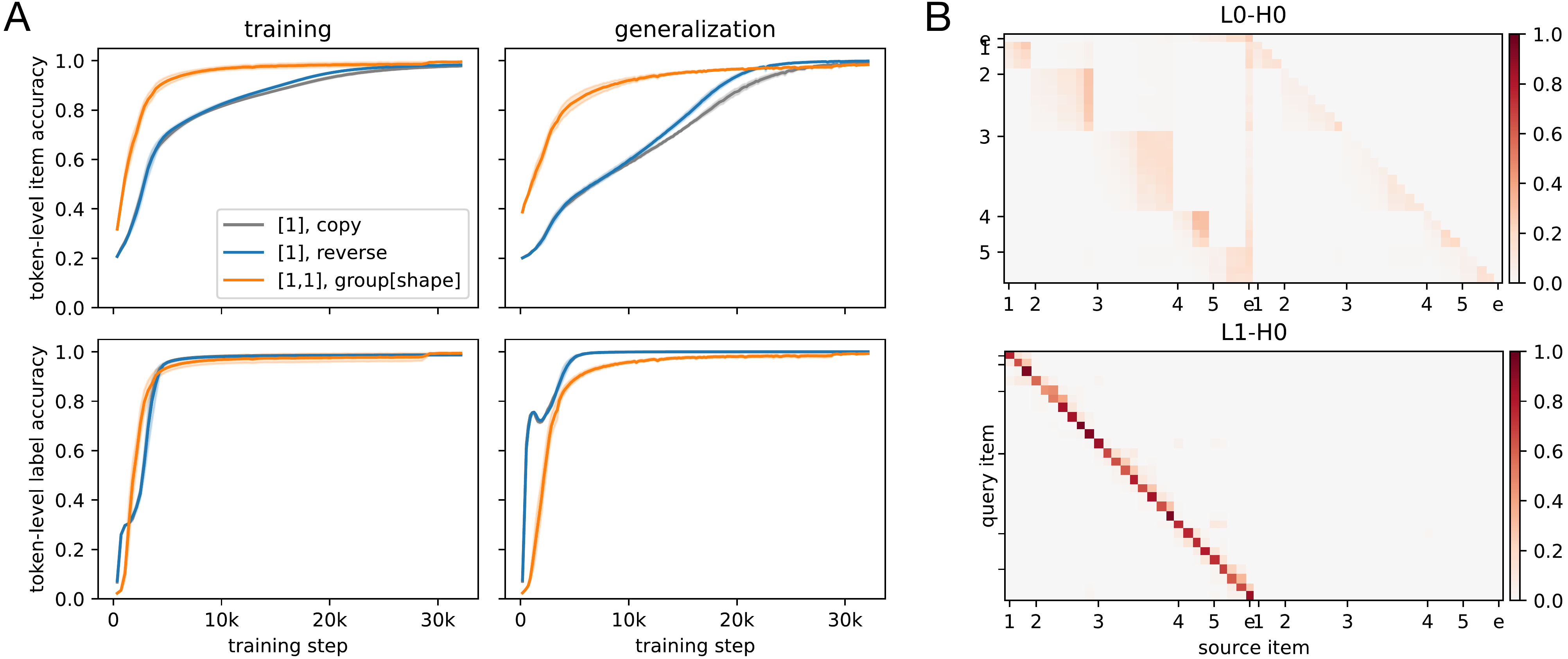}
    \caption{ 
    Learning other tasks in the single-task setting. 
    \textbf{A}. Single-layer, single-headed models can learn the \textsc{copy} task and the \textsc{reverse} task, while two-layer, single-headed models learn the \textsc{group[shape]} task. 
    \textbf{B}. The attention maps from the \textsc{group[shape]} model resemble that from the \textsc{sort[shape]} model (visualization as in Fig~\ref{fig:single-task-attn}A).
    }
    \label{fig:st-append-other}
\end{figure}

\begin{figure}[h]
    \centering
    \includegraphics[width=14cm]{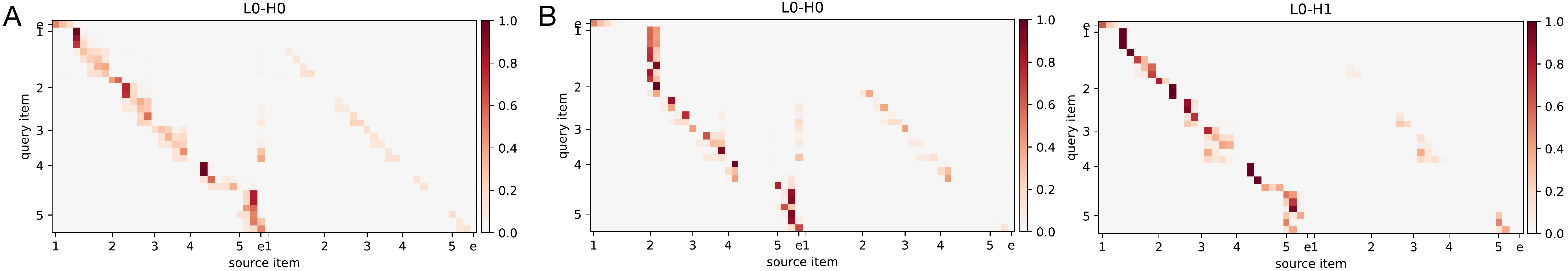}
    \caption{ 
    Attention maps from single-layer models in an example generalization sequence.
    \textbf{A}. Single-layer, single-headed model. 
    \textbf{B}. Single-layer, two-headed model. Visualization as in Fig~\ref{fig:single-task-attn}A.
    }
    \label{fig:st-append-attn}
\end{figure}

\FloatBarrier

\subsection{Multi-task learning}
\label{sec:mt-append}

\begin{figure}[h]
    \centering
    \includegraphics[width=14cm]{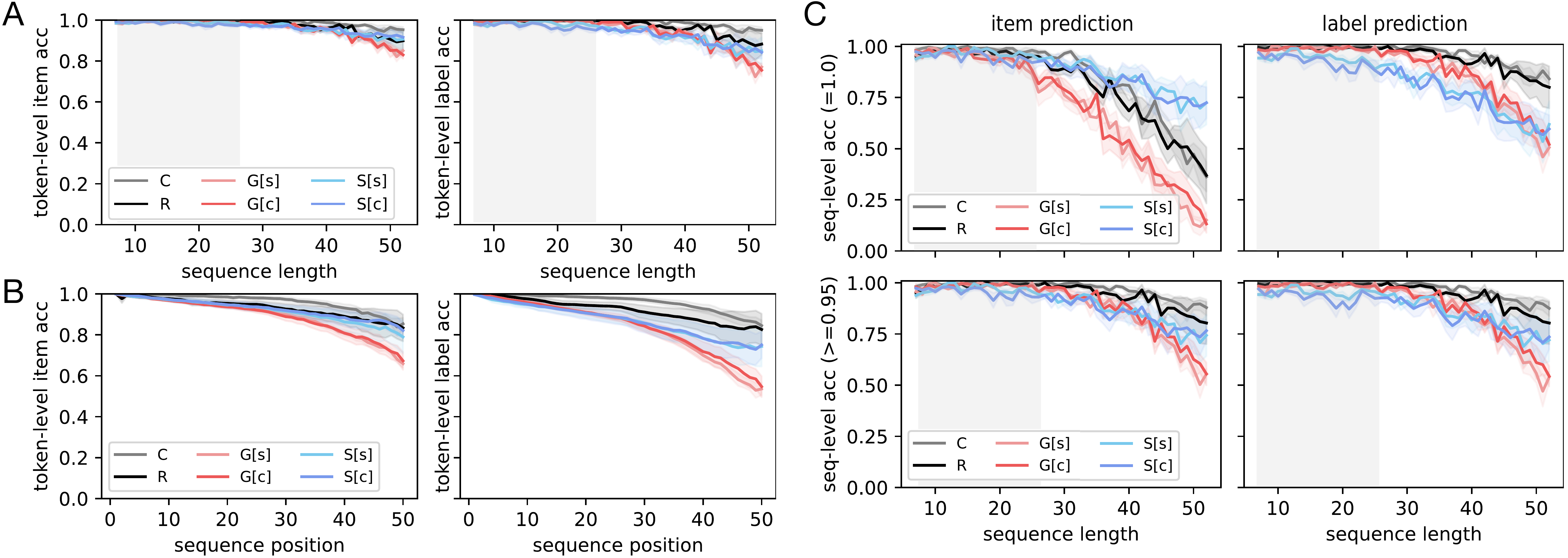}
    \caption{
    Visualization as in Fig~\ref{fig:multi-task-len}, but item and label predictions were obtained using top1 rollout.
    }
    \label{fig:mt-append-len}
\end{figure}

\begin{figure}[h]
    \centering
    \includegraphics[width=9cm]{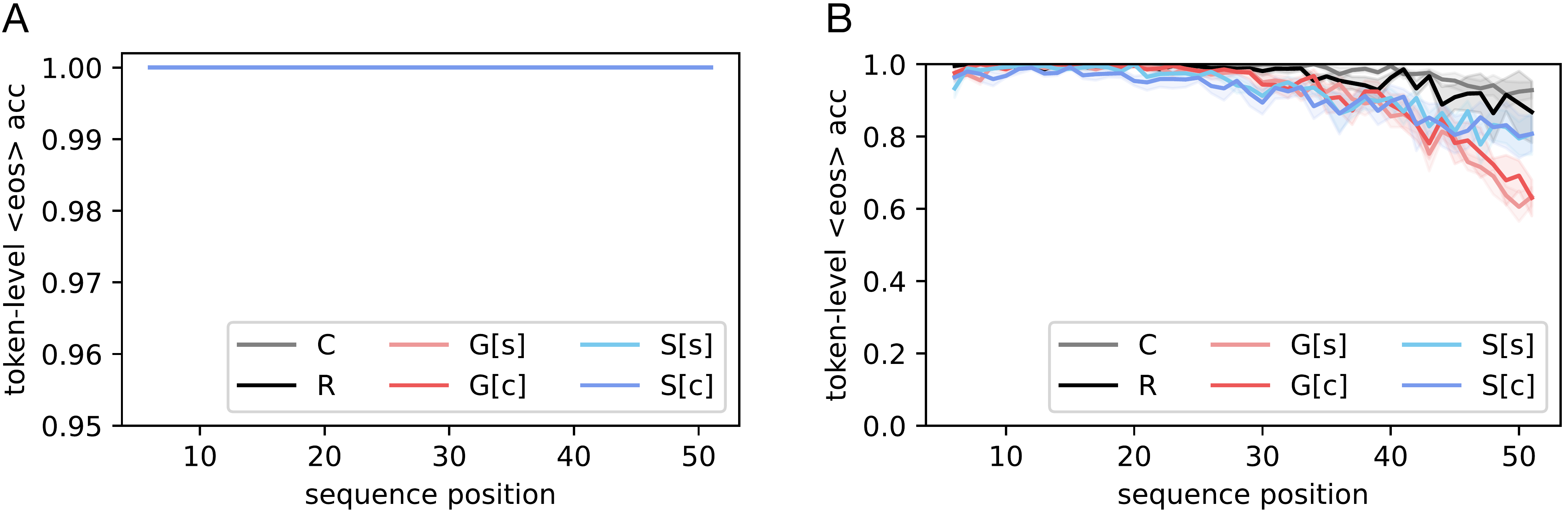}
    \caption{
    \textbf{A}. Six-task models predict the \textsc{eos} token perfectly under teacher forcing.
    \textbf{B}. Prediction of the \textsc{eos} token deteriorates under top1 rollout in six-task models.
    Results were taken from 1k training sequences and 1k length generalization sequences across the five top-performing runs.
    }
    \label{fig:mt-append-eos}
\end{figure}

\begin{figure}[h]
    \centering
    \includegraphics[width=14cm]{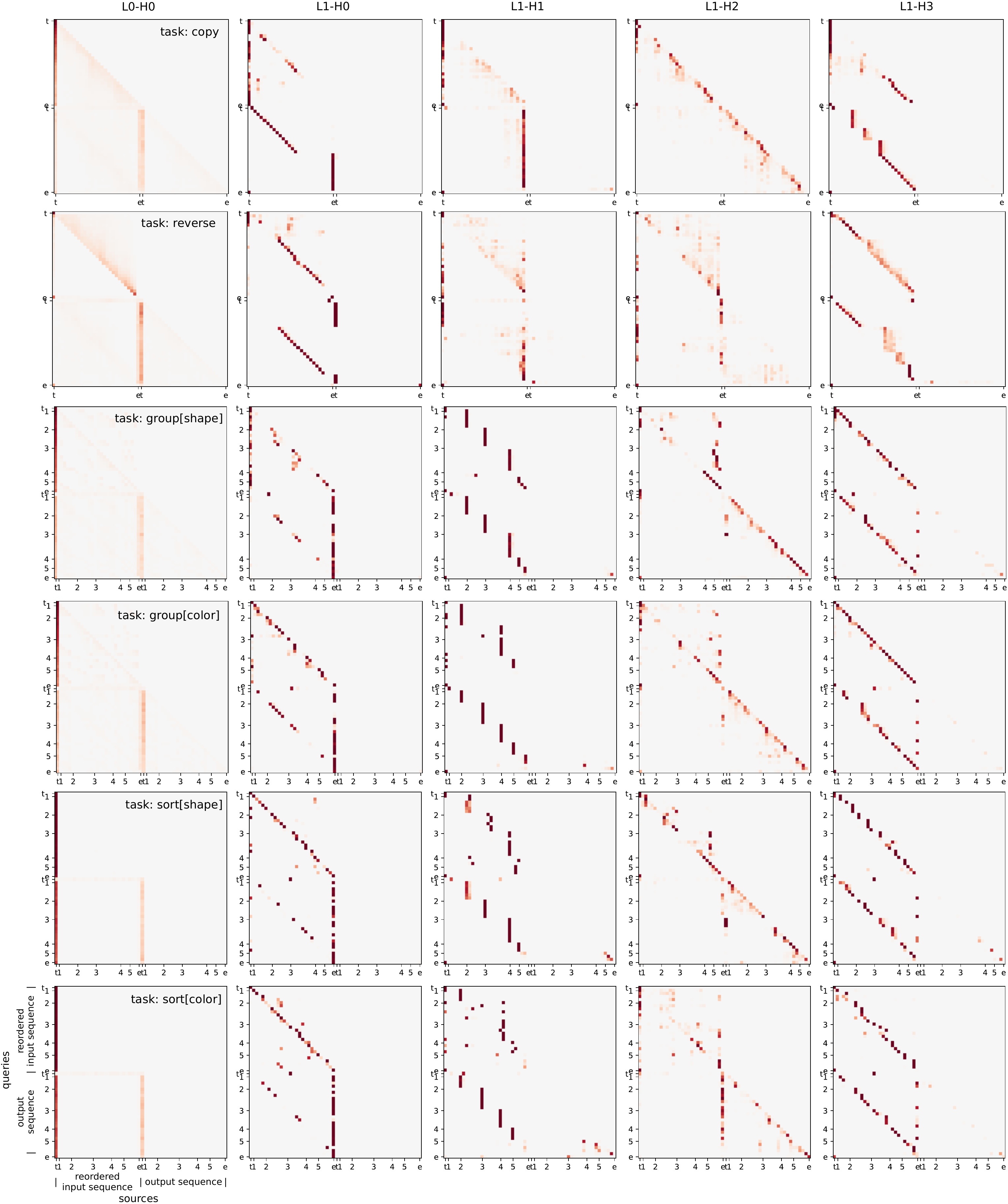}
    \caption{
    Attention maps for an example generalization sequence (for the top-performing multi-task model with [1,4] architecture).  Each row shows the attention maps for one task, and each column corresponds to a single attention head denoted by the layer and head index.  Tokens corresponding to the input sequence were reorderd to match their order in the output sequence for visualization purposes.  Label t indicates the task token.  Label e indicates the \textsc{eos} token.  In the bottom four rows, number labels ranging from 1-5 mark the beginning of the items in each first-level feature group (shape or color). 
    }
    \label{fig:mt-append-attn-maps}
\end{figure}

\begin{figure}[h]
    \centering
    \includegraphics[width=14cm]{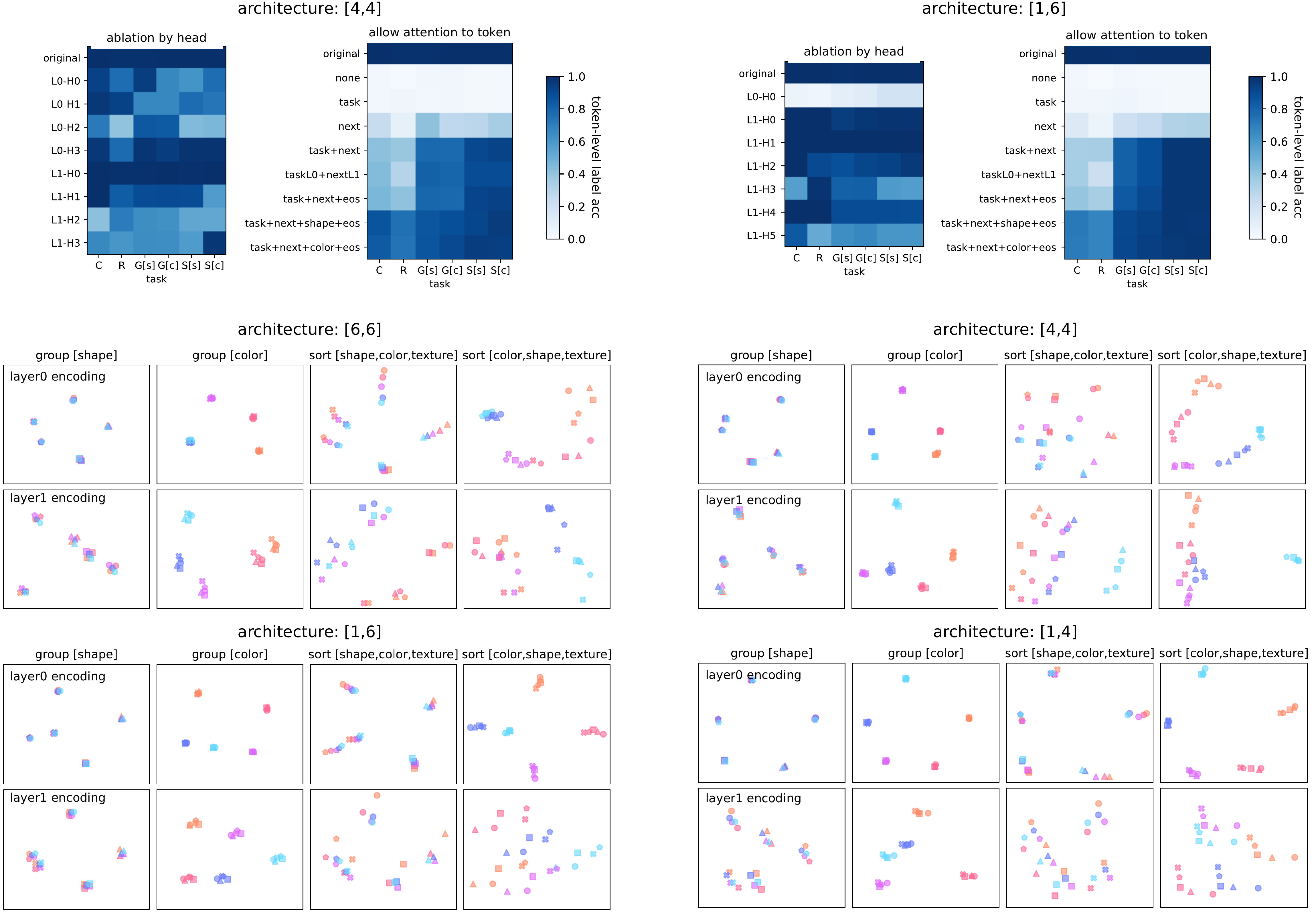}
    \caption{
    Attention ablation and task-conditioned item representations in other top-performing models.  Visualization as in Fig~\ref{fig:multi-task-embed-attn}B and Fig~\ref{fig:pca}.
    }
    \label{fig:mt-append-attn-pca}
\end{figure}

\FloatBarrier

\section{Hyperparameters}
\label{sec:hyperparam}

\begin{table}[h]
\caption{Model and experiment hyperparameters.}
\label{table:hparam}
\centering
\begin{tabular}{|c | c | c|} 
 \hline
 \textbf{Hyperparameter} & \textbf{Single-task learning} & \textbf{Six-task learning} \\
 \hline
 Number of layers & 1 or 2 & 2 \\ 
 \hline
 Number of attention heads & \multicolumn{2}{c|}{(see paper)} \\
 \hline
 Embedding dimension & 128 (two-layer) or 184 (single-layer) & 192 \\ 
 \hline
 MLP hidden layer size & \multicolumn{2}{c|}{64} \\ 
 \hline
 Activation function & \multicolumn{2}{c|}{ReLU} \\
 \hline
 Batch size & \multicolumn{2}{c|}{128} \\ 
 \hline
 Training teacher forcing rate & \multicolumn{2}{c|}{1.0}  \\ 
 \hline
 Optimizer & \multicolumn{2}{c|}{Adam} \\ 
 \hline
 Learning rate & $10^{-4}$ & $5\cdot10^{-4}$  \\ 
 \hline
\end{tabular}
\end{table}

\section{Quantitative Performance}
\label{sec:quant-result}

\begin{table}[h]
\caption{Token-level item and label accuracy across 54k generalization sequences in single-task models.  Mean and standard deviation across four random seeds are shown for each architecture.}
\label{table:single-task-accuracy}
\centering
\begin{tabular}{|c|c|c|c|c|}
 \hline
 task & architecture & position encoding & item prediction & label prediction \\
 \hline
 \textsc{C} & [1] & label & 99.03$\pm$0.41 & 100.00$\pm$0.00 \\
 \hline
 \textsc{R} & [1] & label & 99.75$\pm$0.21 & 100.00$\pm$0.00 \\
 \hline
 \textsc{S[s]} & [1] & label & 76.49$\pm$2.99 & 51.75$\pm$5.35 \\
 \hline
 \textsc{S[s]} & [2] & label & 91.01$\pm$2.76 & 78.02$\pm$5.89 \\
 \hline
 \textsc{S[s]} & [1,1] & label & 99.60$\pm$0.07 & 98.41$\pm$0.23 \\
 \hline
 \textsc{G[s]} & [1,1] & label & 98.41$\pm$0.32 & 99.31$\pm$0.16 \\
 \hline
 \textsc{S[s]} & [1,1] & sinusoidal & 57.04$\pm$10.75	 & 16.50$\pm$6.52 \\
 \hline
 \textsc{S[s]} & [1,1] & learnable & 73.15$\pm$11.65 & 40.51$\pm$14.28 \\
 \hline
\end{tabular}
\end{table}

\begin{table}[h]
\caption{Token-level prediction accuracy across 75k generalization sequences in six-task models (including 500 unique sequences for each length evaluated within each task).  Mean and standard deviation across four random seeds are shown for each architecture. Model architectures are denoted by [\# heads in the first layer, \# heads in the second layer].}
\label{table:multi-task-accuracy}
\centering

\begin{subtable}[h]{\textwidth}
\caption{Attention-balanced models.}
\label{tab:balanced-acc}
\centering
    \begin{tabular}{|c|c|c|c|c|c|c|}
    \hline
    \multicolumn{2}{|c|}{Tasks} & [1,1] & [2,2] & [3,3] & [4,4] & [6,6] \\
    \hline
    \multirow{2}{*}{all} & item & 83.38$\pm$13.05 & 97.04$\pm$0.88 & 98.55$\pm$0.46 & 98.79$\pm$0.94 & \textbf{99.11$\pm$0.32} \\
    & label & 87.21$\pm$7.86 & 96.46$\pm$1.24 & 97.98$\pm$0.42 & 98.68$\pm$0.84 & \textbf{99.17$\pm$0.32} \\
    \hline
    \multirow{2}{*}{\textsc{C}} & item & 89.28$\pm$14.44 & 97.57$\pm$1.19 & 98.81$\pm$0.77 & 98.98$\pm$0.95 & \textbf{99.13$\pm$0.97} \\
    & label & 97.39$\pm$0.84 & 98.59$\pm$0.99 & 99.46$\pm$0.39 & 99.64$\pm$0.34 & \textbf{99.79$\pm$0.17} \\
    \hline
    \multirow{2}{*}{\textsc{R}} & item & 64.30$\pm$24.46 & 96.95$\pm$1.21 & 98.95$\pm$0.45 & 98.69$\pm$0.94 & \textbf{99.20$\pm$0.60} \\
    & label & 94.56$\pm$8.26 & 98.53$\pm$0.80 & 99.64$\pm$0.16 & 99.39$\pm$0.29 & \textbf{99.73$\pm$0.19} \\
    \hline
    \multirow{2}{*}{\textsc{G[s]}} & item & 86.47$\pm$13.51 & 96.51$\pm$0.93 & 98.42$\pm$0.66 & 98.52$\pm$1.15 & \textbf{98.54$\pm$0.47} \\
    & label & 91.54$\pm$8.30 & 97.83$\pm$1.07 & 99.07$\pm$0.33 & \textbf{99.23$\pm$0.52} & 99.05$\pm$0.30 \\
    \hline
    \multirow{2}{*}{\textsc{G[c]}} & item & 86.30$\pm$13.39 & 96.72$\pm$0.80 & 98.28$\pm$0.65 & 98.45$\pm$1.13 & \textbf{98.62$\pm$0.47} \\
    & label & 91.69$\pm$7.79 & 97.93$\pm$1.06 & 98.98$\pm$0.19 & 99.21$\pm$0.46 & \textbf{99.22$\pm$0.24} \\
    \hline
    \multirow{2}{*}{\textsc{S[s]}} & item & 86.97$\pm$8.80 & 97.28$\pm$1.00 & 98.48$\pm$0.29 & 99.01$\pm$0.82 & \textbf{99.59$\pm$0.25} \\
    & label & 74.75$\pm$11.47 & 93.21$\pm$2.23 & 95.53$\pm$1.01 & 97.09$\pm$1.99 & \textbf{98.62$\pm$0.82} \\
    \hline
    \multirow{2}{*}{\textsc{S[c]}} & item & 86.77$\pm$8.63 & 97.20$\pm$0.95 & 98.37$\pm$0.33 & 99.08$\pm$0.70 & \textbf{99.62$\pm$0.13} \\
    & label & 73.57$\pm$10.59 & 92.73$\pm$2.05 & 95.23$\pm$0.90 & 97.53$\pm$1.71 & \textbf{98.65$\pm$0.54} \\
    \hline
    \end{tabular}
\end{subtable}

\vspace{1 em}

\begin{subtable}[h]{\textwidth}
\caption{Attention-frontload models.}
\label{tab:frontload-acc}
\centering
    \begin{tabular}{|c|c|c|c|c|c|}
    \hline
    \multicolumn{2}{|c|}{Tasks} & [2,1] & [3,1] & [4,1] & [6,1] \\
    \hline
    \multirow{2}{*}{all} & item & 69.45$\pm$17.65 & 68.05$\pm$19.27 & 77.86$\pm$22.15 & \textbf{97.76$\pm$0.16} \\
    & label & 79.95$\pm$8.53 & 78.35$\pm$11.60 & 86.53$\pm$11.67 & \textbf{97.52$\pm$0.31} \\
    \hline
    \multirow{2}{*}{\textsc{C}} & item & 57.93$\pm$27.09 & 54.07$\pm$30.36 & 72.01$\pm$30.48 & \textbf{98.50$\pm$0.81} \\
    & label & 97.66$\pm$1.64 & 95.18$\pm$3.28 & 98.48$\pm$1.10 & \textbf{99.61$\pm$0.22} \\
    \hline
    \multirow{2}{*}{\textsc{R}} & item & 57.97$\pm$26.94 & 55.33$\pm$29.52 & 70.49$\pm$32.68 & \textbf{98.78$\pm$0.46} \\
    & label & 95.74$\pm$3.34 & 94.90$\pm$3.50 & 97.97$\pm$2.17 & \textbf{99.56$\pm$0.23} \\
    \hline
    \multirow{2}{*}{\textsc{G[s]}} & item & 70.16$\pm$16.79 & 71.29$\pm$16.08 & 79.20$\pm$19.34 & \textbf{96.85$\pm$0.23} \\
    & label & 80.56$\pm$11.79 & 79.37$\pm$12.51 & 87.82$\pm$12.52 & \textbf{98.96$\pm$0.26} \\
    \hline
    \multirow{2}{*}{\textsc{G[c]}} & item & 70.33$\pm$16.54 & 70.99$\pm$16.21 & 78.86$\pm$19.18 & \textbf{96.50$\pm$0.47} \\
    & label & 79.18$\pm$12.71 & 78.71$\pm$12.82 & 87.28$\pm$12.69 & \textbf{98.82$\pm$0.32} \\
    \hline
    \multirow{2}{*}{\textsc{S[s]}} & item & 79.94$\pm$11.69 & 78.00$\pm$12.49 & 83.99$\pm$15.22 & \textbf{98.03$\pm$0.23} \\
    & label & 62.76$\pm$13.85 & 61.17$\pm$19.50 & 76.69$\pm$18.13 & \textbf{94.17$\pm$0.74} \\
    \hline
    \multirow{2}{*}{\textsc{S[c]}} & item & 80.03$\pm$11.15 & 78.22$\pm$12.21 & 82.46$\pm$16.50 & \textbf{97.95$\pm$0.29} \\
    & label & 64.28$\pm$12.87 & 61.27$\pm$18.82 & 71.27$\pm$24.19 & \textbf{94.08$\pm$0.82} \\
    \hline
    \end{tabular}
\end{subtable}

\vspace{1 em}

\begin{subtable}[h]{\textwidth}
\caption{Attention-backload models.}
\label{tab:2c}
\centering
    \begin{tabular}{|c|c|c|c|c|c|}
    \hline
    \multicolumn{2}{|c|}{Tasks} & [1,2] & [1,3] & [1,4] & [1,6] \\
    \hline
    \multirow{2}{*}{all} & item & 96.06$\pm$0.24 & 98.65$\pm$0.80 & \textbf{99.40$\pm$0.29} & 99.29$\pm$0.17 \\
    & label & 95.48$\pm$0.30 & 97.37$\pm$0.89 & \textbf{99.21$\pm$0.43} & 99.12$\pm$0.36 \\
    \hline
    \multirow{2}{*}{\textsc{C}} & item & 96.86$\pm$0.57 & 99.10$\pm$0.74 & \textbf{99.46$\pm$0.34} & 99.22$\pm$0.52 \\
    & label & 97.78$\pm$0.71 & 99.62$\pm$0.24 & \textbf{99.85$\pm$0.08} & 99.59$\pm$0.55 \\
    \hline
    \multirow{2}{*}{\textsc{R}} & item & 95.99$\pm$0.59 & 98.82$\pm$1.08 & 99.47$\pm$0.37 & \textbf{99.56$\pm$0.14} \\
    & label & 98.09$\pm$0.52 & 99.55$\pm$0.39 & \textbf{99.83$\pm$0.04} & 99.82$\pm$0.09 \\
    \hline
    \multirow{2}{*}{\textsc{G[s]}} & item & 96.12$\pm$0.38 & 98.28$\pm$1.00 & \textbf{98.98$\pm$0.43} & 98.96$\pm$0.16 \\
    & label & 97.70$\pm$0.42 & 98.77$\pm$0.66 & \textbf{99.29$\pm$0.27} & 99.25$\pm$0.07 \\
    \hline
    \multirow{2}{*}{\textsc{G[c]}} & item & 95.82$\pm$0.72 & 98.53$\pm$0.76 & \textbf{99.17$\pm$0.34} & 98.78$\pm$0.40 \\
    & label & 97.51$\pm$0.51 & 99.06$\pm$0.42 & \textbf{99.51$\pm$0.18} & 99.13$\pm$0.32 \\
    \hline
    \multirow{2}{*}{\textsc{S[s]}} & item & 95.86$\pm$0.77 & 98.51$\pm$0.72 & \textbf{99.68$\pm$0.26} & 99.59$\pm$0.13 \\
    & label & 91.12$\pm$0.83 & 93.54$\pm$2.44 & 98.40$\pm$1.16 & \textbf{98.41$\pm$0.89} \\
    \hline
    \multirow{2}{*}{\textsc{S[c]}} & item & 95.72$\pm$0.54 & 98.70$\pm$0.71 & \textbf{99.64$\pm$0.28} & 99.63$\pm$0.29 \\
    & label & 90.74$\pm$0.66 & 93.76$\pm$2.15 & 98.36$\pm$1.05 & \textbf{98.54$\pm$1.10} \\
    \hline
    \end{tabular}
\end{subtable}
\hfill

\end{table}

\end{document}